%% file: main.tex
\documentclass[]{xbench}
\makeatletter
\newcommand\blfootnote[1]{%
  \begingroup
  \renewcommand\thefootnote{}\footnote{#1}%
  \addtocounter{footnote}{-1}%
  \endgroup
}
\makeatother
\input{sections/00_decl}

\title{AgentIF-OneDay: A Task-level Instruction-Following Benchmark for General AI Agents in Daily Scenarios}
\author{xbench.org}
\xbenchdata[Code]{\texttt{\href{https://github.com/xbench-ai}{github.com/xbench-ai}}}
\xbenchdata[Data]{\texttt{\href{https://huggingface.co/xbench-ai}{huggingface.co/xbench-ai}}}
\xbenchdata[Correspondence]{team@xbench.org}
\usepackage{xcolor}
\definecolor{ogreen}{rgb}{0,0.6,0}

\abstract{
\input{sections/0_abstract}
}

\begin{document}
\maketitle
\blfootnote{* Equal contribution.} 

\input{sections/1_introduction}

\input{sections/2_related}
\input{sections/3_what_is_taskif}
\input{sections/4_how_to_build_taskif}

\input{sections/5_experiments}
\input{sections/6_discussion}
\input{sections/7_conclusion}

\input{sections/authors}

\bibliography{iclr2025_conference}
\bibliographystyle{iclr2025_conference}
\input{sections/appendix}
\end{document}

%% file: sections/00_decl.tex
\usepackage[utf8]{inputenc}
\usepackage{subcaption}

\usepackage[T1]{fontenc}    %
\usepackage{hyperref}       %
\usepackage{url}            %
\usepackage{booktabs}       %
\usepackage{amsfonts}       %
\usepackage{nicefrac}       %
\usepackage{microtype}      %
\usepackage[dvipsnames]{xcolor}

\usepackage{nicematrix}
\usepackage{algorithm}
\usepackage{algpseudocode}
\usepackage{amsthm}
\usepackage[normalem]{ulem}
\usepackage{wrapfig}
\usepackage{graphicx}
\usepackage{cancel}
\usepackage{multirow}
\usepackage{multicol}
\usepackage{amssymb}
\usepackage{cleveref}

\usepackage[colorinlistoftodos]{todonotes}

\crefformat{section}{\S#2#1#3}
\Crefformat{section}{\S#2#1#3}
\crefmultiformat{section}{\S#2#1#3}{ and \S#2#1#3}{, \S#2#1#3}{ and \S#2#1#3}
\Crefmultiformat{section}{\S#2#1#3}{ and \S#2#1#3}{, \S#2#1#3}{ and \S#2#1#3}
\crefrangeformat{section}{\S#3#1#4 to \S#5#2#6}
\Crefrangeformat{section}{\S#3#1#4 to \S#5#2#6}

\usepackage{colortbl}
\usepackage{xspace}
\usepackage{inconsolata}

\usepackage{graphicx}

\usepackage{xcolor,ifthen,marginnote}
\newboolean{showcomments}

\usepackage{etoolbox}
\newtoggle{release}
\toggletrue{release}

\usepackage{siunitx}

\usepackage{placeins}

\usepackage{makecell}
\usepackage{siunitx}

\tcbset{
  aibox/.style={
    width=\linewidth,
    top=8pt,
    bottom=4pt,
    colback=black!5!white,
    colframe=ogreen,
    colbacktitle=ogreen,
    enhanced,
    center,
    attach boxed title to top left={yshift=-0.1in,xshift=0.15in},
    boxed title style={boxrule=0pt,colframe=white,},
  }
}
\newtcolorbox{AIbox}[2][]{aibox,title=#2,#1}

\usepackage{listings}
\usepackage{tcolorbox} 

%% file: sections/0_abstract.tex
The capacity of AI agents to effectively handle tasks of increasing duration and complexity continues to grow, demonstrating exceptional performance in coding, deep research, and complex problem-solving evaluations. However, in daily scenarios, the perception of these advanced AI capabilities among general users remains limited. We argue that current evaluations prioritize increasing task difficulty without sufficiently addressing the diversity of agentic tasks necessary to cover the daily work, life, and learning activities of a broad demographic.

To address this, we propose \method, aimed at determining whether general users can utilize natural language instructions and AI agents to complete a diverse array of daily tasks. These tasks require not only solving problems through dialogue but also understanding various attachment types and delivering tangible file-based results.

The benchmark is structured around three user-centric categories: \textbf{Open Workflow Execution}, which assesses adherence to explicit and complex workflows; \textbf{Latent Instruction}, which requires agents to infer implicit instructions from attachments; and \textbf{Iterative Refinement}, which involves modifying or expanding upon ongoing work. 
We employ instance-level rubrics and a refined evaluation pipeline that aligns LLM-based verification with human judgment, achieving an 80.1\% agreement rate using Gemini-3-Pro. \method comprises 104 tasks covering 767 scoring points. 

We benchmarked four leading general AI agents and found that agent products built based on APIs and ChatGPT agents based on agent RL remain in the first tier simultaneously. Leading LLM APIs and open-source models have internalized agentic capabilities, enabling AI application teams to develop cutting-edge Agent products. 

%% file: sections/1_introduction.tex
\section{Introduction}
AI agents driven by Large Language Models (LLMs) have recently exhibited remarkable progress in autonomous reasoning, planning, and tool usage. Although agents have achieved success in handling increasingly complex tasks within vertical domains like vibe coding~\citep{jimenez2023swe,zan2025multi}, deep research~\citep{wei2025browsecomp,chen2025xbench} and complex problem solving~\citep{phan2025humanity}, these advancements are not yet fully palpable to ordinary users. We posit that current evaluations lack the necessary diversity to reflect real-world scenarios across work, life, and learning.

Addressing this gap, we propose \method, a framework to assess the efficacy of AI agents in assisting general users with daily tasks through language instructions. \method demands the delivery, analysis, and modification of comprehensive work outputs, structured around three key dimensions:

\begin{enumerate}
    \item \textbf{Open Workflow Execution:} These tasks assess the agent's ability to execute automated processes precisely and exhaustively when the user provides a clear and detailed operational procedure. This tests the agent's robustness in long-context processing, avoidance of "instruction forgetting," and suppression of hallucinations.
    \item \textbf{Latent Instruction Inference:} This dimension requires the agent to autonomously induce and infer implicit, unstated general rules or constraints from task attachments and apply them to new tasks.
    \item \textbf{Iterative Refinement:} These tasks simulate a multi-turn, progressive collaborative process between the user and the agent. The agent must perform precise content updates and adjustments based on supplementary or corrective instructions from the user upon existing outputs, measuring the efficacy of its state maintenance and human-machine collaboration.
\end{enumerate}

We control the distribution of topics in the evaluation set to balance \textit{alignment with authentic user needs} and \textit{maximization of task diversity}. The former includes common demands such as daily office software usage, life planning, and research, helping us understand the correlation between agent capabilities and real-world value. The latter primarily increases the difficulty of "hack" evaluation metrics through out-of-domain tasks, including game strategy research and the automation of scientific research workflows.

Beyond manually annotated tasks, we designed a \textbf{File-centered Automated Agentic Pipeline} for task generation. We utilize the ChatGPT agent to collect information-dense attachments with high potential for question generation. Using human-annotated tasks as seeds, we extract and augment workflow frameworks. Finally, we synthesize the attachments and workflows to produce \method evaluation items that are logically coherent and deeply correlated with the attachment information. Furthermore, this methodology is extensible to broader agentic data synthesis applications.

We employ \textbf{Instance-level rubrics} for evaluation and have enhanced the consistency between LLM-based Verifiers and human scoring through pipeline design. For instance, we implemented visual parsing for PPT and HTML files and utilized Vision-Language Models (VLMs) as verifiers. Additionally, for rubrics requiring real-time verification, the model defaults to a Search Mode. The utilization of superior multimodal foundation models has significantly improved the accuracy of the discriminator. When employing Gemini-3-Pro as the judge model, the consistency between \method scoring points and human scoring reaches 80.6\%.

\method comprises \textbf{104} tasks and covers \textbf{767} scoring points. Through systematic testing and analysis of mainstream AI agent products on this benchmark, we aim to reveal the strengths and bottlenecks of current agent products and agentic models, providing empirically valuable references for the future optimization of AI agents. Simultaneously, the high-quality, finely annotated instruction data accumulated by \method holds significant potential as training data for reinforcement learning.

%% file: sections/2_related.tex
\section{Related Work}
\textbf{Instruction Following Benchmarks.} 
Instruction following benchmarks are crucial for evaluating the ability of large language models to execute tasks based on specific instructions. For instance, IFEval \citep{zhou2023instructionfollowingevaluationlargelanguage} assesses a model's capability using a set of verifiable instructions to ensure accuracy. Another notable benchmark, FOFO \citep{xia2024fofobenchmarkevaluatellms}, classifies tasks into various domains and imposes specific format requirements, such as the Prescription Format Standard, to test a model's precision. To further push the boundaries of evaluation, ComplextBench \citep{wen2024benchmarkingcomplexinstructionfollowingmultiple} generates instructions through a compositional approach, which is specifically designed to enhance the assessment of a model's understanding of more complex commands.

\textbf{Agent Benchmarks.}
Agent benchmarks are designed to evaluate the ability of AI agents to perform complex, real-world tasks that often require a combination of skills \citep{liu2023agentbench,hu2024infiagent}. GAIA \citep{mialon2023gaiabenchmarkgeneralai}, for example, assesses an agent's reasoning and tool-using capabilities by having it complete a variety of real-world tasks. Similarly, VisualWebArena \citep{koh2024visualwebarenaevaluatingmultimodalagents} focuses on an agent's ability to execute tasks specifically within a web environment, which is a common and important domain. Additionally, tau-Bench \citep{yao2024taubenchbenchmarktoolagentuserinteraction} provides another dedicated framework for evaluating a model's capacity for effective tool usage.

\textbf{Evaluation on Real-World Economic Tasks.}
Recent research has shifted focus from general abilities to evaluating AI's potential for economic labor automation.
\citet{handa2025economictasksperformedai} analyzed millions of user conversations, revealing that AI usage is predominantly concentrated in software development and writing tasks, with 57\% of interactions focusing on augmenting human capabilities rather than full automation.
Similarly, \citet{chatterji2025people} examined consumer usage patterns of ChatGPT, finding that while non-work interactions are growing, the model provides significant economic value as a decision-support tool for knowledge-intensive tasks.
To rigorously assess professional capabilities, \citet{patwardhan2025gdpvalevaluatingaimodel} introduced GDPval, a benchmark evaluating AI on long-horizon, multi-modal tasks across 44 occupations.
Complementing this, \citet{mazeika2025remotelaborindexmeasuring} proposed the Remote Labor Index (RLI) based on real-world freelance projects, finding that current frontier agents achieve only a 2.5\% automation rate on these complex, end-to-end workflows.
Focusing on the growth trajectory of these capabilities, \citet{kwa2025measuringaiabilitycomplete} introduced the ``50\%-task-completion time horizon'' metric, observing that this capability horizon doubles approximately every seven months, projecting that AI could automate month-long software tasks within five years.

\textbf{Reinforcement for Instruction Following.}
Reinforcement learning has emerged as a powerful paradigm for training models to follow complex instructions by leveraging feedback signals. VerIF \citep{peng2025verifverificationengineeringreinforcement} uses both soft and hard verifiable signals to provide direct rewards for correctly executed instructions. \citet{ren2025tradeoffselfsupervisedreinforcementlearning} employs self-supervised RL to tackle multi-constraint instructions. This technique works by decomposing complex instructions into simpler ones with an incrementally increasing number of constraints, which in turn generates preference pairs for instruction following.

%% file: sections/3_what_is_taskif.tex
\section{\method}
Current evaluations of instruction following predominantly focus on individual models, chatbots, or the isolated agentic capabilities of foundation models. Conversely, assessments targeting complete agent systems tend to be confined to vertical domains. However, as general-purpose agent products undergo rapid iteration, both their capability boundaries and user expectations are evolving dynamically.

To address this, we introduce \method, a benchmark characterized by the following core features:

\begin{enumerate}
    \item \textbf{End-to-End Task Completion Assessment:} Each task involves not only multimodal input/output (text and images) but also includes attachments that mirror real-world requirements. We impose no constraints on the architecture of the evaluated system; instead, the primary criterion for success is whether the delivered outcome strictly adheres to the instructions.
    
    \item \textbf{Alignment with Real-World General Agent Usage and Long-Tail Coverage:} \method is domain-agnostic. We collected authentic requirements from a broad spectrum of general agent users, encompassing daily office routines, personal hobbies, and specialized professional tasks. This approach ensures that evaluation scores effectively align with user value, while the inclusion of long-tail tasks mitigates the risk of overfitting or ``gaming'' the benchmark. Furthermore, capabilities such as image editing and in-depth research—now common in chatbot applications—are included if they constitute the task deliverable. Crucially, we do not prescribe the method of solution; for instance, image editing may be accomplished via software manipulation rather than solely relying on text-to-image generation models.
    
    \item \textbf{Instance-Level Rubrics with Workflow-Based Judging:} Evaluating \method requires the judge to comprehend diverse file formats and utilize tools for factual verification. The advancement of multimodal foundation models has significantly enhanced the efficacy of automated judges. We anticipate that Workflow Judges and Judge Agents will become increasingly prevalent standard practices in future evaluations.
\end{enumerate}

\input{sections/3_1_category}
\input{sections/3_2_eval}
\input{sections/3_3_staticstics}

%% file: sections/3_1_category.tex
\subsection{Task Categories}
\begin{figure}
    \centering
    \includegraphics[width=\linewidth]{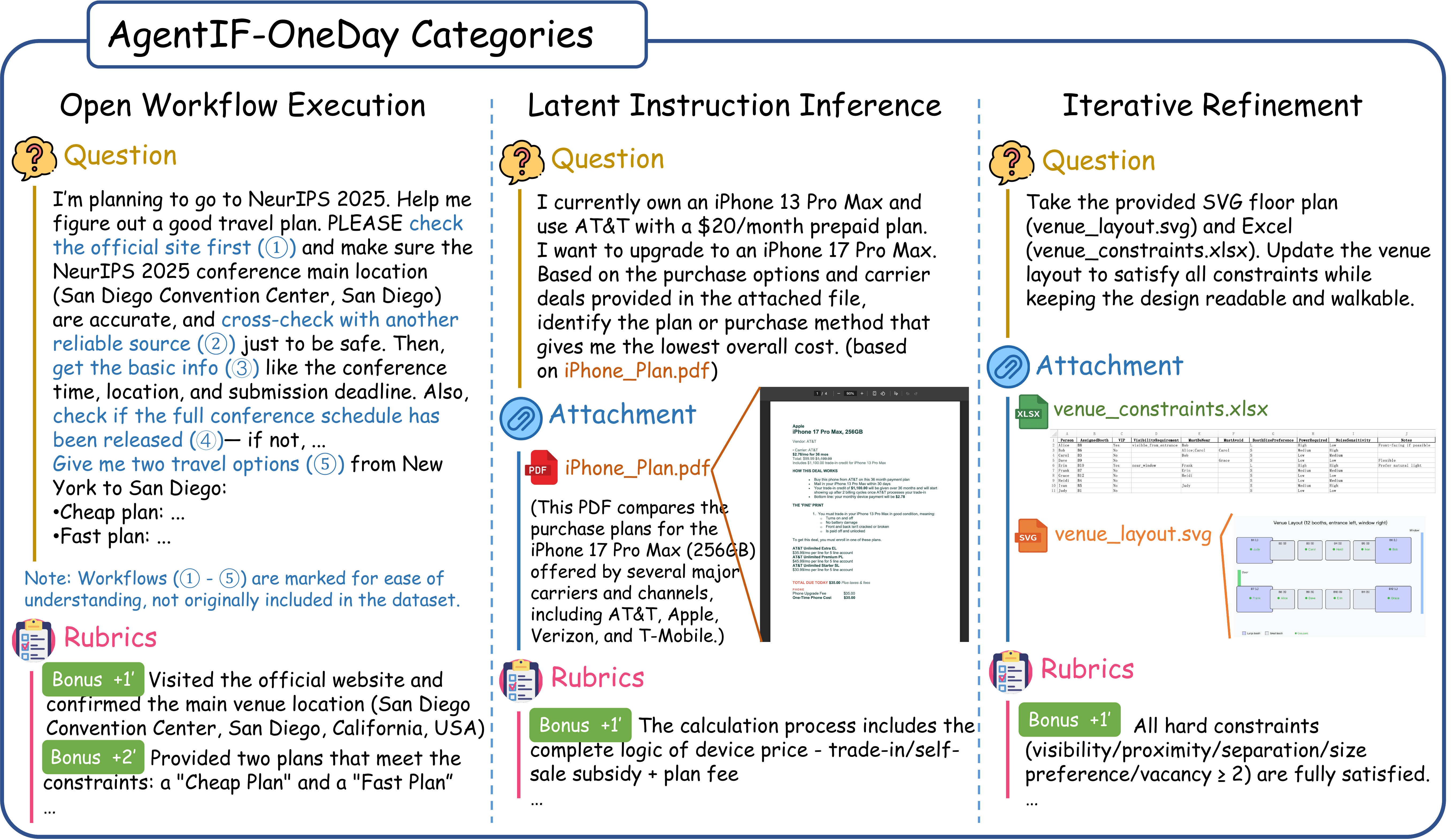}
    \caption{\method categorizes tasks into three types based on how users interact with agent products.}
    \label{fig:taskif_main}
\end{figure}

We categorize instructions into three types based on how users interact with general AI agents: Open Workflow Execution, Latent Instruction Inference, and Iterative Refinement. We introduce each type in sequence and present example questions in \cref{fig:taskif_main}.

\textbf{Open Workflow Execution} evaluates an agent's ability to automate tasks precisely and exhaustively when presented with explicit, detailed operational procedures. The core competency lies in managing long-context information and strictly adhering to instruction sequences, requiring the agent to overcome ``instruction forgetting'' or hallucinations to faithfully reconstruct user-defined steps. For instance, in the ``NeurIPS 2025 Travel Planning'' task, the user dictates a rigorous five-step workflow rather than a simple query. The agent must first verify the San Diego Convention Center via the official website and cross-reference with third-party sources. Subsequently, after confirming dates and submission deadlines, it must check for the detailed schedule before finally generating ``economical'' and ``fast'' travel options from New York based on this verified data. This process demands not only information retrieval but also a faithful execution of the ``verify-then-plan'' logical chain, ensuring every output is grounded in the precise execution of preceding steps.

Distinct from explicit instruction following, \textbf{Latent Instruction Inference} (or context-based conditional reasoning) focuses on an agent's ability to mine implicit logic from provided materials and generalize it to new tasks. In real-world scenarios, users often provide reference documents or case studies rather than explicit formulas, requiring the agent to ``comprehend'' and apply latent constraints. For example, in the ``iPhone 17 Pro Max Purchase Decision'' task, the user provides only their current device model (iPhone 13 Pro Max) and carrier status, requesting the scheme with the ``lowest total cost.'' The agent cannot derive the answer directly from the prompt; it must analyze the attached \texttt{iPhone\_Plan.pdf} to deduce complex pricing logic—integrating the base device price, trade-in values for specific models, and varying carrier plan costs. The agent must accurately identify the trade-in value of the user's current device from the table and perform calculations based on the plan duration. This task rigorously tests the agent's depth of understanding of unstructured information and its reasoning capabilities under implicit constraints.

\textbf{Iterative Refinement} simulates the incremental modification scenarios common in human-machine collaboration, assessing the agent's ability to refine and optimize existing outputs based on specific constraints and feedback. Rather than generating content from scratch, the agent must perform operations such as adjustment, recalculation, and layout optimization while maintaining the current state, which demands strong state consistency management. In the example task shown, the user provides a floor plan file (\texttt{venue\_layout.svg}) and a constraints file (\texttt{venue\_constraints.xlsx}). The agent faces a complex optimization challenge: it must update the venue layout to satisfy all hard constraints listed in the Excel file---such as visibility, proximity, separation, size preference, and vacancy requirements---while keeping the design readable and walkable. Success depends on the agent's ability to process multi-modal inputs (SVG and Excel), accurately interpret specific constraints, and seamlessly integrate these rules into the existing layout to deliver a valid and functional final design.

%% file: sections/3_2_eval.tex
\subsection{Evaluation}

\method employs an instance-level, rubric-based scoring mechanism that assigns specific verification checks to the conditions and tasks within each problem. This system is characterized by three key properties:

(1) \textit{Binary Scoring}, where each rubric item is strictly evaluated as satisfied or not to ensure objectivity and mitigate LLM judging bias;

(2) \textit{Distinction between Bonus and Penalty Items}, separating the assessment of capability (bonus items for key requirements) from error rates (penalty items for critical mistakes), reflecting the distinct tolerance levels for system capabilities versus failures; 

(3) \textit{File-Content Alignment}, enabling the direct evaluation of agent-generated files alongside textual output. The final scoring aggregates these components. 

Let $N$ be the total number of problems. For the $i$-th problem, let $S^+_i$ represent the sum of satisfied bonus points and $S^-_i$ the sum of triggered penalty points. The maximum achievable score for the problem is denoted as $S^{max}_i$. The normalized score for problem $i$, denoted as $s_i$, is calculated by clamping the net score at zero and normalizing against the maximum:
\begin{equation}
    s_i = \frac{\max(0, S^+_i - S^-_i)}{S^{max}_i}
\end{equation}
The final \method score is the mean of these normalized scores across all problems:
\begin{equation}
    \text{Score}_{\text{final}} = \frac{1}{N} \sum_{i=1}^{N} s_i
\end{equation}

\begin{figure}
    \centering
    \includegraphics[width=0.5\linewidth]{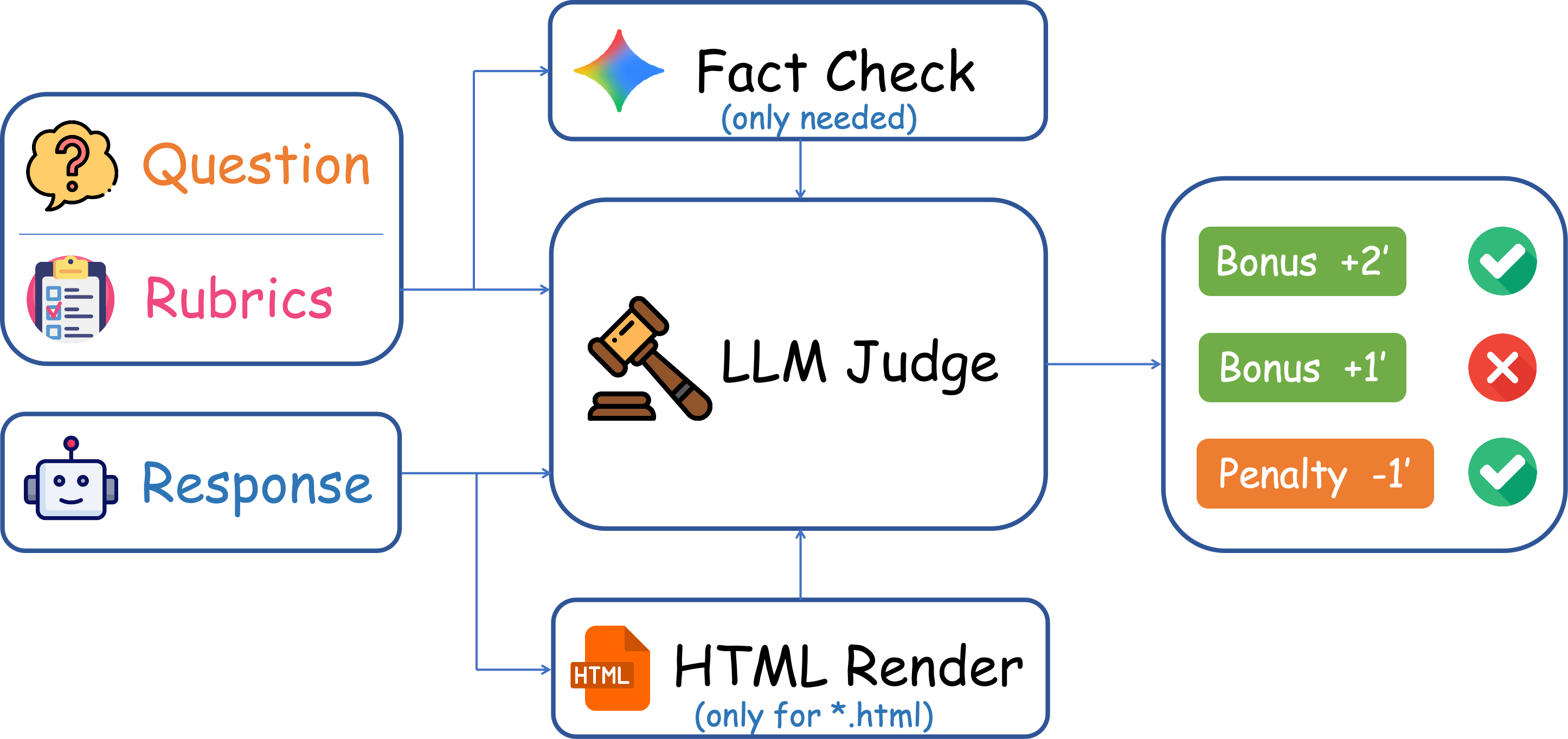}
    \caption{Evaluation pipeline in \method.}
    \label{fig:evaluation_pipeline}
\end{figure}

As shown in Fig~\ref{fig:evaluation_pipeline}, we adopt an LLM-as-judge approach, leveraging the rubrics provided in each task for scoring. For each rubric, the LLM only determines whether it is satisfied, and the final score is computed by summing or subtracting the points associated with the corresponding rubrics and performing normalization.

Since some tasks rely on facts that may change over time, we utilize web search for factual verification. Specifically, for each task, the LLM decomposes it into facts that need verification, which are then grounded with Google Search \footnote{\url{https://ai.google.dev/gemini-api/docs/google-search}}. For tasks that require writing HTML code, we render the HTML to provide a clear visual perception, thus avoiding the need for LLMs to directly read complex HTML code to perform scoring. Prompts for evaluation are provided in the Appendix~\ref{app:llm_as_judge}.

%% file: sections/3_3_staticstics.tex
\subsection{\method Statistics}
\begin{figure*}[t] 
    \centering
    \includegraphics[width=\linewidth]{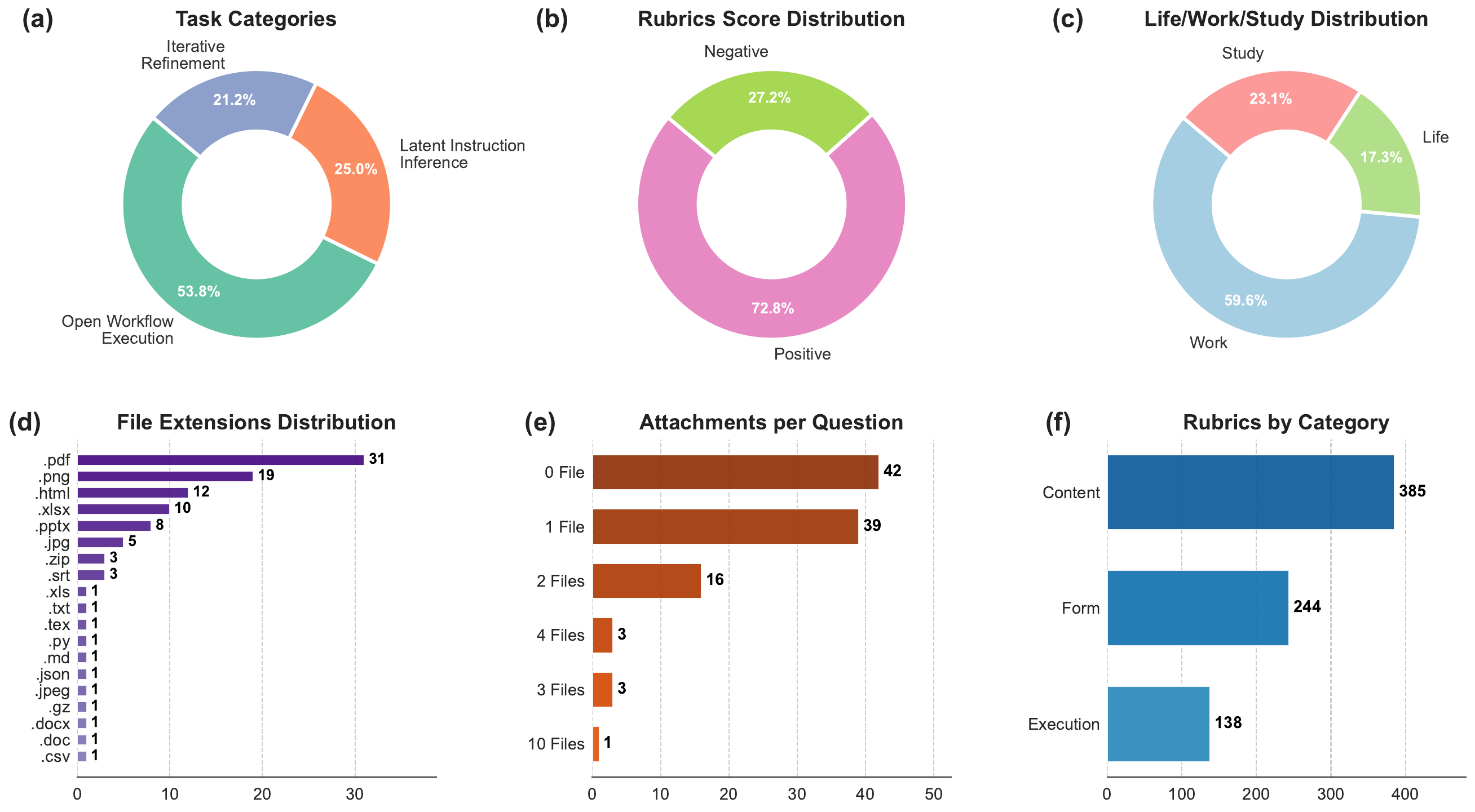}
    \caption{Statistical overview of \method: (a) task category distribution, (b) rubric score polarity, (c) domain distribution, (d) file extension frequency, (e) attachment count per task, and (f) rubric categories.}
    \label{fig:taskif_stats}
\end{figure*}

Figure \ref{fig:taskif_stats} presents a comprehensive statistical overview of the \method benchmark. As shown in the task category distribution (Figure \ref{fig:taskif_stats}(a)), the workflow is dominated by \textit{Open Workflow Execution} (\textbf{53.8\%}), followed by \textit{Latent Instruction Inference} (\textbf{25.0\%}) and \textit{Iterative Refinement} (\textbf{21.2\%}). The domain distribution (Figure \ref{fig:taskif_stats}(c)) illustrates that the tasks are primarily centered around \textit{Work} scenarios (\textbf{59.6\%}), with significant representation in \textit{Study} (\textbf{23.1\%}) and \textit{Life} (\textbf{17.3\%}) domains, ensuring a balance between professional, academic, and personal user needs.

Reflecting the complexity of the tasks, Figure \ref{fig:taskif_stats}(e) shows the distribution of attachments per question. While a plurality of tasks involve \textbf{0 Files} (42 tasks) or \textbf{1 File} (39 tasks), a notable subset requires handling multiple attachments, with some tasks involving up to \textbf{10 files}, testing multi-file reasoning capabilities. As detailed in Figure \ref{fig:taskif_stats}(d), the dataset encompasses a wide variety of file extensions: \textit{.pdf} (31) and \textit{.png} (19) are the most frequent, but the inclusion of diverse formats such as code files (.html, .py), structured data (.xlsx, .csv), and media (.jpg, .srt) rigorously tests the agent's multimodal tool utilization.

To ensure granular evaluation, the benchmark utilizes a robust rubric system. Figure \ref{fig:taskif_stats}(b) shows the rubric score distribution, where the majority of criteria are \textit{Positive} (\textbf{72.8\%}), focusing on successful task completion, while \textbf{27.2\%} are \textit{Negative}, likely testing for constraints and safety adherence. The categorical breakdown in Figure \ref{fig:taskif_stats}(f) highlights the specific areas of evaluation: \textit{Content} is the most heavily weighted category with \textbf{385} points, followed by \textit{Form} (\textbf{244} points) and \textit{Execution} (\textbf{138} points), ensuring a holistic assessment of the agent's output quality and procedural accuracy.

%% file: sections/4_how_to_build_taskif.tex
\input{sections/4_1_collection}
\input{sections/4_2_synthetic_data}

%% file: sections/4_1_collection.tex
\section{Building \method}
\subsection{Collection}
We developed a detailed annotation guideline for collecting instances from human labelers. Each labeler was asked to submit original questions within their verified area of expertise, following detailed guidelines emphasizing that questions should be: (1) Difficult: not answerable by non-experts, (2) Objective: with a single, unambiguous correct answer that domain experts would agree upon, and (3) Search resistant: with correct answers not readily retrievable via standard search engines.

Submitted questions underwent a multi-stage review pipeline:
(1) Initial screening: Editorial review for formatting, clarity, and adherence to guidelines;
(2) Expert validation: Review by at least one additional domain expert (distinct from the original author) who attempted to answer the question and provided feedback on its quality;
(3) Revision: Author revisions based on feedback from the validators to improve clarity or appropriate difficulty;
(4) Final review: A quality check by an independent editorial team before inclusion.
Questions were rejected at any stage if they were found to be ambiguous, too easy (i.e., answerable via a simple web search), or lacking a clear, correct answer.

%% file: sections/4_2_synthetic_data.tex
\subsection{Data Synthesis and Rewriting}

\begin{figure}
    \centering
    \includegraphics[width=\linewidth]{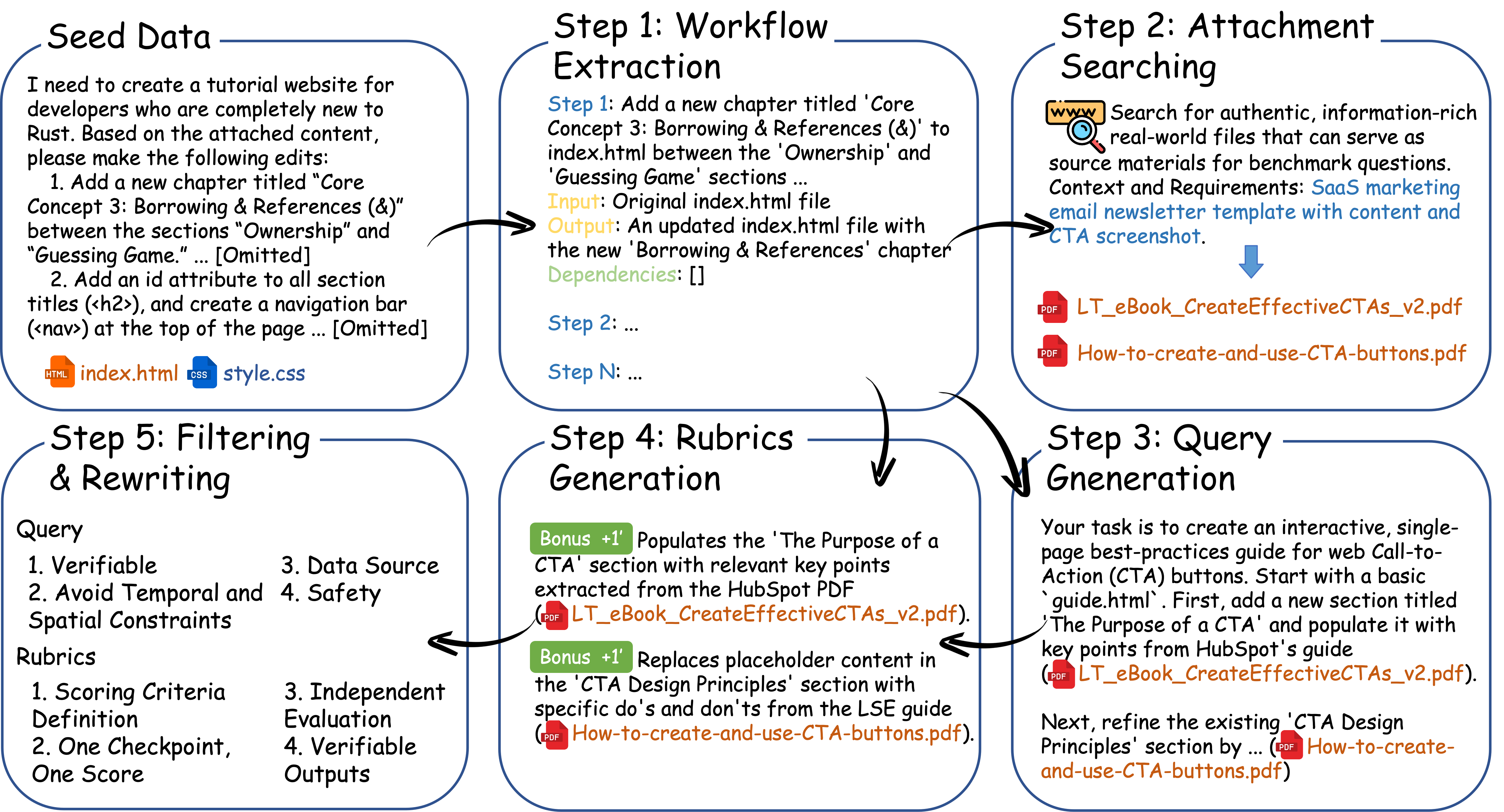}
    \caption{Workflow of synthesizing instances in \method.}
    \label{fig:taskif_synthesize}
\end{figure}

We designed an automatic synthetic data generation pipeline to expand the dataset by leveraging high-quality, human-authored seed tasks, but
reduce cost, as shown in Fig.~\ref{fig:taskif_synthesize}. This process automatically generates new problems and corresponding evaluation criteria that share the same workflow as the seed tasks but are instantiated in different application scenarios. The synthesized problems, attachments, and rubrics undergo a rigorous human review and refinement process. The final synthetic tasks are evaluated using the same scoring methodology as the human-authored tasks. We used Gemini-2.5-Pro during the synthesizing.

The pipeline operates as a multi-stage process anchored by the extracted workflow driven by diverse human-authored seed tasks. This workflow acts as the structural backbone, which eventually transforms single human-authored task into a diverse set of synthetic tasks combined with attachments. The detailed prompts for each module are provided in Appendix~\ref{app:synthetic_data_generation}.

\textbf{Step 1: Workflow Extraction.} The process begins by analyzing a high-quality human seed task to isolate the underlying logic. LLMs are instructed to analyze the seed task to extract the logical sequence of operations required for completion, including the steps, required inputs, expected outputs, and dependencies between steps. This abstract workflow serves as the structural backbone for all subsequent generations.

\textbf{Step 2: Attachment Searching.} To ensure that the synthetic tasks are grounded in realistic contexts, the pipeline procures and analyzes domain-specific supplementary materials. Based on the functional requirements of the extracted workflow, diverse search queries (e.g., detailed financial report dashboard or medical invoice template) are generated to retrieve relevant documents and images from the web with Search mode in ChatGPT \footnote{All web content was accessed in November 2025.}. The retrieved attachments are then analyzed by ChatGPT, which summarizess the content, identifies key data points, and determines the suitability of the files for specific task types, ensuring that the visual data align with the intended logic.

\textbf{Step 3: Query Generation.} Using the fixed workflow template and the analyzed attachments, the pipeline generates specific task instructions. New task instructions are generated to preserve the original logical workflow while introducing new content, domains, and contexts. Crucially, specialized prompts are employed to guide the LLM in producing tasks that align with the specific characteristics of the three core categories, ensuring that the synthetic data preserves the cognitive demands of the original human-designed categories. 

\textbf{Step 4: Rubrics Generation.}
Rubrics are formulated by strictly following human annotation guidelines to construct Bonus Criteria and Penalty Criteria. The rubric adheres to the principles of ``one-score-per-point'' and independent verification.

\textbf{Step 5: Filtering and Rewriting.} To ensure the automatically generated tasks meet the high standards of \method, we implement a strict filtering and manual refinement process. We specifically filter for complex tasks with more than 3 steps and apply the following optimization criteria:  
(1) \textit{Answer Verifiability}: The task must yield a clear, measurable answer (e.g., a specific numerical value, a generated file, or a defined output string). Vague requirements (e.g., ``Analyze market trends'') are discarded or concretized (e.g., ``Calculate the year-over-year growth rate'');  
(2) \textit{Spatio-Temporal Independence}: Instructions relying on relative time or location (e.g., ``last week,'' ``nearby'') are replaced with fixed data ranges, specific timestamps, or absolute locations (e.g., ``Analyze stock changes from Jan--Jun 2024'') to ensure the task remains valid over time;  
(3) \textit{Data Source Validity}: All referenced URLs, documents, or data sources are verified for existence and relevance. Tasks relying on non-existent resources are either fixed with reliable alternatives or discarded;  
(4) \textit{Security and Ethics}: Instructions requiring sensitive operations—such as account registration, logging in, or accepting cookie policies—are removed to prevent security risks and platform dependencies;  
(5) \textit{Rubric Alignment}: The generated rubrics are manually reviewed to ensure the distinction between Bonus and Penalty items is clear. Evaluation emphasizes final, verifiable artifacts, while intermediate actions (e.g., successfully visiting a website) are not scored.

%% file: sections/5_experiments.tex
\section{Experiments}
\subsection{Setup}
We selected four advanced and popular agents for testing to comprehensively evaluate their performance in complex tasks and diverse scenarios. All agent platforms were accessed and configured in December 2025. These agents represent leading generative AI technology and Agent design paradigms currently available on the market: (1) \textbf{ChatGPT-Agent}\footnote{\url{https://chatgpt.com/}}, (2) \textbf{Genspark}\footnote{\url{https://www.genspark.ai/}}, (3) \textbf{Manus}\footnote{Manus 1.5, \url{https://manus.im/}}, and (4) \textbf{Minimax-Agent}\footnote{Pro mode, \url{https://agent.minimaxi.com/}}. We selected \textit{Gemini-3-Pro-preview} as the judge model, with \texttt{max\_tokens} set to 100{,}000 and \texttt{temperature} set to 0.1.

\subsection{Results}
Table~\ref{tab:agent_performance} presents the main results of our evaluation across the four selected agents. The performance is gauged by the overall success rate across all tasks. The top performing agent is \textbf{Manus}, achieving an overall score of $0.645$. \textbf{Genspark} and \textbf{ChatGPT-Agent} follow closely, with scores of $0.635$ and $0.626$, respectively, forming a cluster of high-performing agents. \textbf{Minimax-Agent} lags behind with $0.562$.

\begin{AIbox}{
\shortstack[l]{ 
    Performance Parity Between Agents Developed by Foundation Model Companies Natively and \\ 
    API-Driven Application Companies
}}
The results demonstrate a crucial performance parity: agents built purely upon leading LLM APIs (relying on prompt engineering and external tooling) achieve success rates comparable to those of custom Agent RL-based systems. This equivalence indicates that the baseline agentic competence has become a commodity, effectively integrated into modern foundational models. Future competitive edge will likely shift from building core agentic skills to optimizing products for specific user needs and leveraging user data to refine the overall agent experience.
\end{AIbox}

\input{tables/5_main}

\subsection{Analysis}

\paragraph{What are the specific strengths of different agents?}

In terms of task domains, ChatGPT stands out as the premier productivity tool, Manus as the ultimate life assistant, and Genspark as the ideal learning partner, shown in \cref{tab:perf_task_domains}.

These three products pursue different iteration paths. The ChatGPT-Agent prioritizes "GDPval" focusing on the user experience within professional work scenarios. In contrast, Manus and Genspark place greater emphasis on user feedback. These distinct evaluation approaches have resulted in different product strengths and weaknesses. We believe that an excellent general-purpose Agent should balance a wide variety of tasks rather than favoring one specific area.

\begin{table}[htbp]
    \centering
    \caption{Agent Performance Ranking in Work, Life, and Study}
    \label{tab:agent-ranking}
    \begin{tabular}{lcccccc}
        \toprule
        \multirow{2}{*}{} & \multicolumn{2}{c}{\textbf{Work}} & \multicolumn{2}{c}{\textbf{Life}} & \multicolumn{2}{c}{\textbf{Study}} \\
        \cmidrule(lr){2-3} \cmidrule(lr){4-5} \cmidrule(lr){6-7}
         & Agent & Score & Agent & Score & Agent & Score \\
        \midrule
        1st & ChatGPT-Agent & 72.18 & Manus & 73.40 & Genspark & 71.19 \\
        \addlinespace 
        2nd & Genspark & 71.86 & ChatGPT-Agent & 69.67 & Manus & 64.41 \\
        \addlinespace 
        3rd & Manus & 70.27 & Genspark & 67.85 & ChatGPT-Agent & 59.29 \\
        \bottomrule
    \end{tabular}
    \label{tab:perf_task_domains}
\end{table}

Regarding capability dimensions, Genspark performs best in implicit instruction inference, Manus excels in open workflow execution, and the Minimax-Agent possesses the best iterative editing capabilities, shown on the right side of \cref{tab:agent_performance}.

The variation in performance across capability dimensions likely stems from differences in Agent frameworks. Implicit condition inference is currently the weakest capability across agents generally. Some tasks require the Agent to automatically identify formatting rules from attachments—such as extracting header/footer structures or citation styles from a PPT template—and apply them to new content generation. We observed that even the best-performing systems struggle to be completely accurate in these tasks. They either replicate the format correctly but with insufficient coverage, or they understand the content well but fail to maintain structural consistency.

\paragraph{How do performance metrics vary across different quality rubrics and attachment conditions?}

Analyzing the breakdown by rubric reveals distinct strengths. \textbf{Genspark} excels at Instruction Following ($0.766$) and handling Negative Constraints ($\mathbf{0.824}$), tying with \textbf{ChatGPT-Agent} in the latter. \textbf{Manus} achieves the highest Factuality score ($0.731$). Intriguingly, \textbf{Minimax-Agent} exhibits the highest performance in Logic/Functionality ($0.755$), suggesting a strong reasoning core despite its lower overall score. Regarding attachment handling, \textbf{Genspark} scores highest with attachments ($0.691$). Notably, \textbf{Manus} maintains virtually identical performance with ($0.646$) and without ($0.644$) attachments, demonstrating exceptional robustness to changes in input modality.

\paragraph{What is the efficiency profile of the tested agents?}

Efficiency, measured by Average Latency (s), presents a clear trade-off with overall quality. \textbf{Genspark} ($484.1$ s) and \textbf{Manus} ($500.0$ s) offer a strong balance between speed and quality. In contrast, \textbf{Minimax-Agent} is the slowest agent, requiring an average of $1416.2$ s, significantly slower than all other agents. This high latency potentially correlates with its specialized strength in Logic/Functionality, suggesting its reasoning process may be more computationally intensive.

\paragraph{Consistency Between Human and LLM Judges.}
We constructed an evaluation set comprising 28 problems and a total of 171 scoring criteria, covering three problem types and various attachment formats. Human annotations were performed on model outputs within the evaluation set, and multiple LLMs were tested for automated scoring.
The performance of different LLM judges in aligning with human annotations is summarized in Table~\ref{tab:llm_judge_accuracy}.

\input{tables/5_judge}

As shown in Table~\ref{tab:llm_judge_accuracy}, \textbf{Gemini-3-Pro-preview} achieved the highest agreement with human judges at $80.1\%$, while \textbf{GPT-5.1} showed an accuracy of $63.8\%$.
The discrepancies observed between different LLM scores are primarily attributable to issues such as hallucination during scoring and inconsistent instruction following. Furthermore, the differences between LLM and human judgments often stem from varying interpretations of abstract concepts, including "conciseness," "relative completeness," and "design sense."
\begin{figure}[htbp]
    \centering
    \begin{subfigure}[b]{0.48\linewidth}
        \centering
        \includegraphics[width=\linewidth]{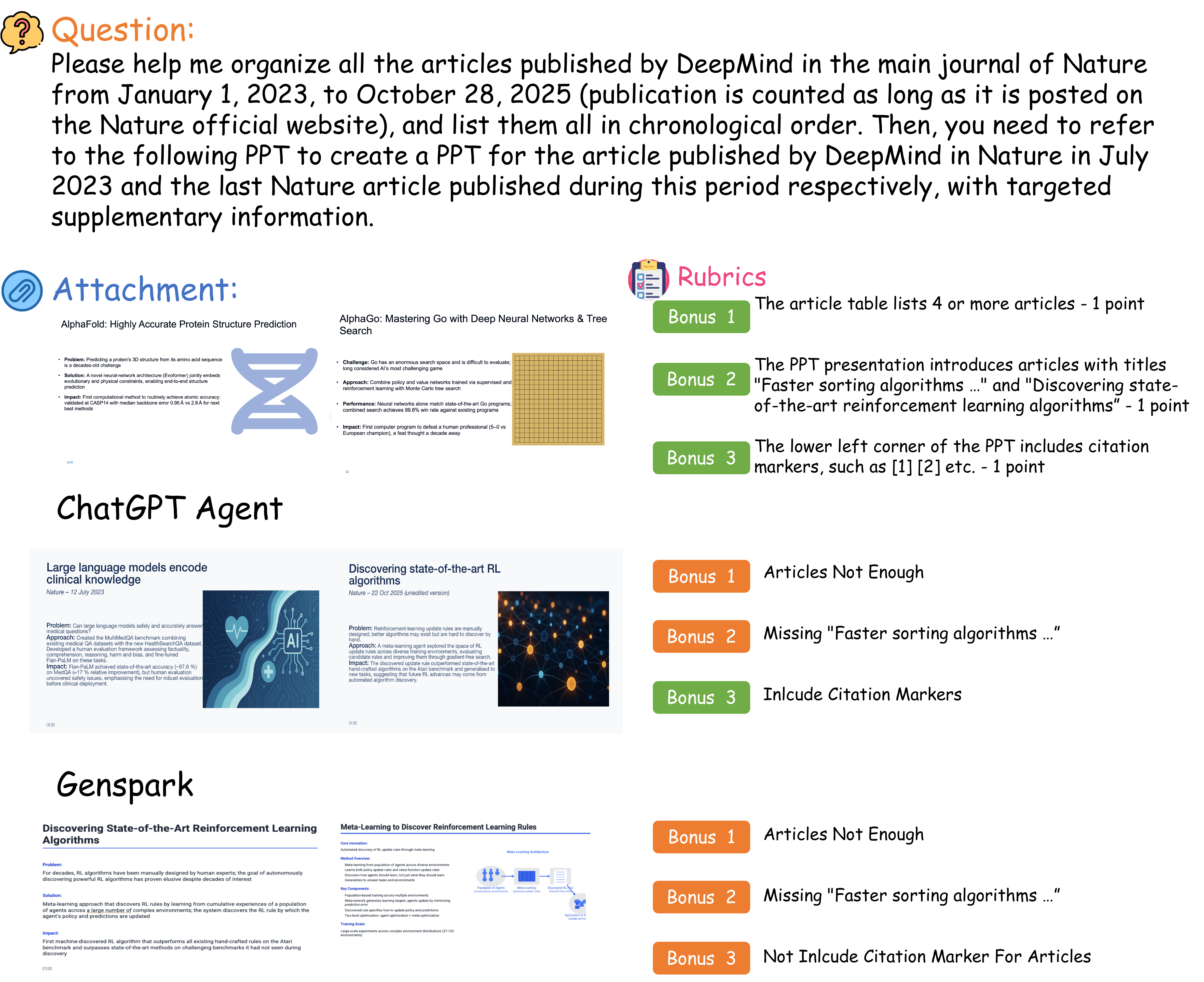}
        \caption{Making a research slide}
        \label{fig:case_study_1}
    \end{subfigure}
    \hfill 
    \begin{subfigure}[b]{0.48\linewidth}
        \centering
        \includegraphics[width=\linewidth]{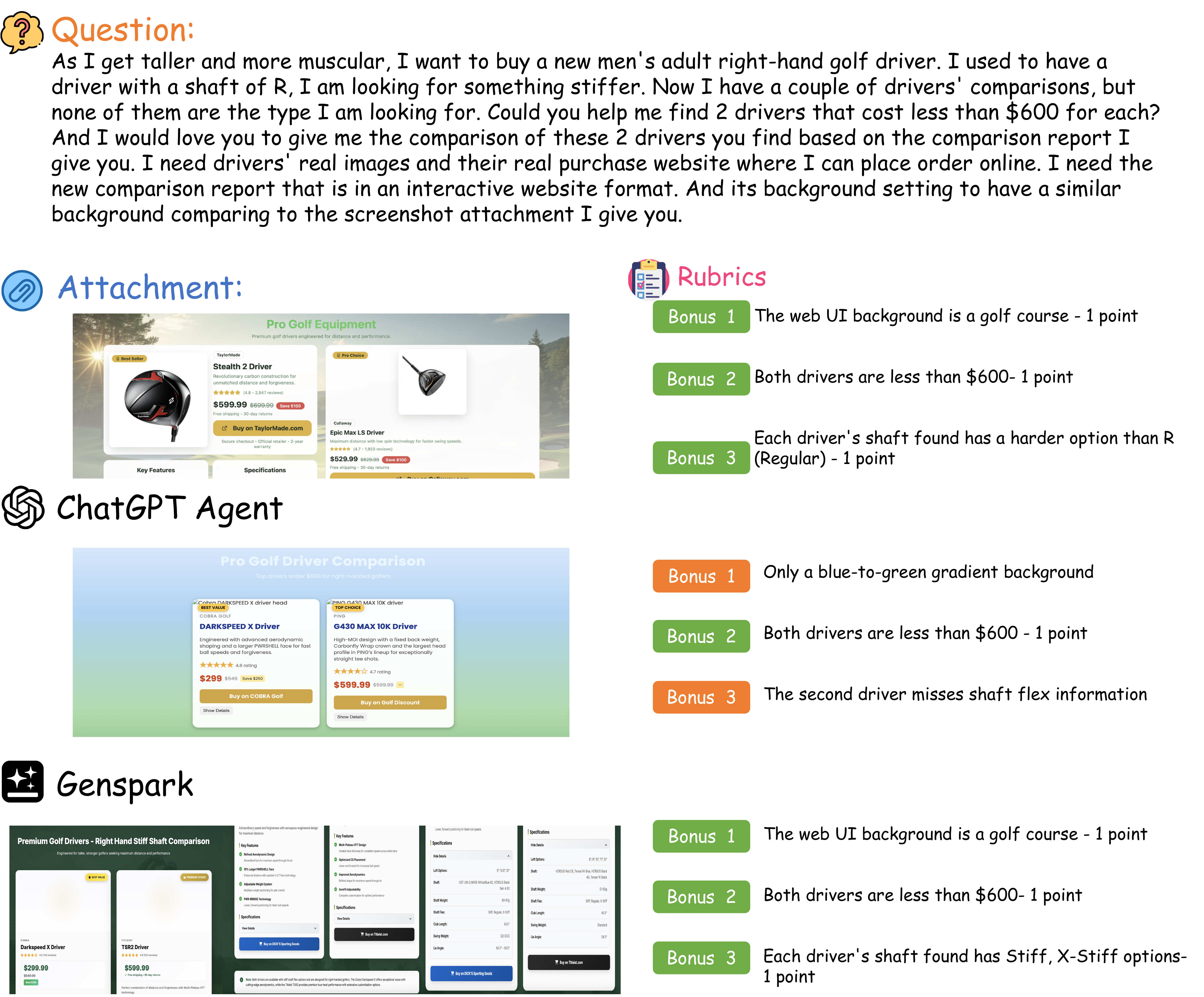} 
        \caption{Cross-platform shopping}
        \label{fig:case_study_2}
    \end{subfigure}
    
    \caption{Case study of latent instruction inference tasks. }
    \label{fig:case_studies_combined}
\end{figure}

\paragraph{Case Study.} We analyze sample problems on latent instruction inference in \cref{fig:case_studies_combined}. The task (a) requires the agent to research articles published by DeepMind in \textit{Nature} within a specified timeframe and organize them into a PowerPoint presentation with a specific format. The format of the PPT is not explicitly described in the text prompt; instead, the agent must understand the format from an attached PPT file. For example, features include each article's introduction slide having an accompanying image and a citation marker located in the bottom left corner. The task requires the agent to mine these intrinsic features from the file and implicitly adhere to them when generating new slides.

It can be observed that the ChatGPT agent followed the page format relatively well, but the number of articles listed was insufficient. Genspark, on the other hand, encountered issues with the PPT format; it missed the citation marker for the second article and included irrelevant information on the third and fourth pages.

In the cross-platform shopping task (b), the user requirement for a background similar to the screenshot compels the agent to perform high-level semantic image understanding and style transfer alongside standard information retrieval. While the ChatGPT agent successfully processes the explicit scalar constraint of a price under six hundred dollars, it demonstrates limitations in multimodal reasoning by disregarding the visual context of the golf course image and misses technical specifications such as shaft stiffness. In contrast, Genspark displays superior cross-modal integration by accurately inferring implicit visual directives to generate context-aware user interfaces while simultaneously satisfying rigorous data filtering requirements, highlighting a significant performance disparity in handling complex bimodal prompts.

Successfully completing this task in full requires the agent to sustain a research effort over a longer duration while maintaining adherence to hidden instructions, which remains a challenge for current agents.

%% file: tables/5_main.tex
\begin{table*}[htbp]
    \centering
    \caption{Agent Performance Summary across Key Metrics.}
    \label{tab:agent_performance}
    \resizebox{\textwidth}{!}{%
        \begin{tabular}{
            l 
            S[table-format=1.3, mode=text] 
            S[table-format=4.1, mode=text] 
            *{9}{S[table-format=1.3, mode=text]} 
        }
            \toprule
            \multirow{2}{*}{\textbf{Agent}} & {\textbf{Overall}} & {\textbf{Avg.}} & \multicolumn{4}{c}{\textbf{Rubric}} & \multicolumn{2}{c}{\textbf{Attachment}} & \multicolumn{3}{c}{\textbf{Instruction Following}} \\
            
            \cmidrule(lr){4-7} \cmidrule(lr){8-9} \cmidrule(lr){10-12}
            
            & {\textbf{Score}} & {\makecell{Latency \\ (s)}} & {\makecell{Inst. \\ Following}} & {\textbf{Factuality}} & {\makecell{Logic/ \\ Func.}} & {\makecell{Neg. \\ Const.}} & {\makecell{w/ Att. \\ Score}} & {\makecell{w/o Att. \\ Score}} & {\makecell{Iter. \\ Ref.}} & {\makecell{Latent \\ Inst.}} & {\makecell{Open \\ Work.}} \\
            \midrule
            
            \textbf{Manus} & \textbf{0.645} & 500.0 & 0.762 & \textbf{0.731} & 0.693 & 0.529 & 0.646 & \textbf{0.644} & 0.646 & 0.610 & \textbf{0.661} \\
            
            \textbf{Genspark} & 0.635 & 484.1 & \textbf{0.766} & 0.663 & 0.720 & \textbf{0.824} & \textbf{0.691} & 0.551 & 0.681 & \textbf{0.719} & 0.577 \\
            
            \textbf{ChatGPT-Agent} & 0.626 & 665.1 & 0.739 & 0.687 & 0.673 & \textbf{0.824} & 0.666 & 0.566 & 0.689 & 0.613 & 0.606 \\
            
            \textbf{Minimax-Agent} & 0.562 & 1416.2 & 0.709 & 0.623 & \textbf{0.755} & 0.588 & 0.603 & 0.502 & \textbf{0.717} & 0.510 & 0.525 \\
            
            
            \bottomrule
        \end{tabular}
    } 
    \vspace{0.1cm}
\end{table*}

%% file: tables/5_judge.tex
\begin{table}[htbp]
    \centering
    \caption{LLM Judge Accuracy Against Human Annotations.}
    \label{tab:llm_judge_accuracy}
    \begin{tabular}{lc}
        \toprule
        \textbf{Judge Model} & \textbf{Consistency} \\
        \midrule
        Gemini-3-Pro-preview & 80.1\% \\
        Gemini-2.5-Pro & 73.9\% \\
        GPT-5.1 & 63.8\% \\
        \bottomrule
    \end{tabular}
\end{table}

%% file: sections/6_discussion.tex
\section{Discussion and Limitation}

\subsection{Challenges in Data Collection}
Constructing the \method presents unique data scaling challenges due to the inherent complexity of the tasks:

\begin{itemize}
    \item \textbf{High Annotation Cost:} Our pilot study indicates that designing a single valid task requires an average of three hours per annotator.
    \item \textbf{Limited Individual Scalability:} Task formulation relies heavily on personal life scenarios and imagination. Annotators quickly exhaust familiar contexts, making continuous high-quality generation difficult. Consequently, scaling requires a broad population of annotators rather than a small group of experts.
    \item \textbf{Verification Heterogeneity:} Unlike vertical domains where background knowledge is shared among experts, daily life scenarios possess highly diverse contexts. It is challenging to recruit experts capable of verifying tasks across such a wide spectrum of topics.
\end{itemize}

To partially address these bottlenecks, we implemented a synthetic generation pipeline. By leveraging models to brainstorm diverse topics for subsequent human refinement, we effectively reduce dependency on individual annotator creativity and specific background knowledge.

\subsection{From OneDay to OneWeek}
Extending agent capabilities to longer time horizons represents the next technical frontier. The design philosophy presented in this paper is transferable from daily to weekly scenarios. 

While recent benchmarks have explored long-horizon tasks (e.g., Game playing~\cite{pleines2025pokemon}, Vending-bench~\cite{backlund2025vending}, and Paper Bench~\cite{starace2025paperbench}), they often focus on specific domains. Although domain-specific training can improve performance on these metrics, it does not guarantee generalization to other tasks of equivalent duration. To accurately assess progress toward Artificial \textbf{General} Intelligence  without relying on extensive domain-specific customization, it remains essential to construct comprehensive ``OneWeek'' benchmarks derived from a broad spectrum of work, learning, and daily life activities.

%% file: sections/7_conclusion.tex
\section{Conclusion}
In this work, we introduced \method, a task-level instruction-following benchmark designed to evaluate general AI agents under realistic, daily-use conditions. Through carefully constructed tasks spanning open workflow execution, latent instruction inference, and iterative refinement, our study highlights that effective agent behavior hinges not merely on raw reasoning ability, but on sustained instruction adherence, file-centered understanding, and stateful collaboration across time. Empirical results across multiple leading agent systems reveal a notable convergence in baseline agentic competence between API-driven and RL-based agents, suggesting that agent-level capabilities are becoming increasingly standardized within modern foundation models. At the same time, the observed performance gaps across task categories and rubric dimensions underscore persistent challenges in implicit constraint inference and long-horizon consistency. Beyond benchmarking, AgentIF-OneDay provides a scalable methodology for generating verifiable, workflow-grounded agent tasks, offering both a diagnostic tool for current systems and a high-quality data source for future agent training. We hope this benchmark will serve as a foundation for evaluating and advancing general-purpose agents toward more reliable, user-aligned assistance in everyday scenarios.

%% file: sections/authors.tex
\newpage
\section{Contributors}

\subsection*{Core Contributors}
Kaiyuan Chen, Qimin Wu, Taiyu Hou, Tianhao Tang, Xueyu Hu, Yuchen Hou

\textit{(These authors contributed equally to this work)}
\subsection*{Contributors}
Bikun Li, Chengming Qian, Guoyin Wang, Haolin Chen, Haotong Tian, Haoye Zhang, Haoyu Bian, Hongbing Pan,
Hongkang Zhang, Hongyi Zhou, Jiaqi Cai, Jiewu Rao, Jiyuan Ren, Keduan Huang, Lucia Zhu Huang,
Mingyu Yuan, Naixu Guo, Qicheng Tang, Qinyan Zhang, Shuai Chen, Siheng Chen, Ting Ting Li, Xiaoxing Guo,
Yaocheng Zuo, Yaoqi Guo, Yinan Wang, Yinzhou Yu, Yize Wang, Yuan Jiang, Yuan Tian,
Yuanshuo Zhang, Yuxuan Liu, Yvette Yan Zeng, Zenyu Shan, Zihan Yin

\subsection*{Supervision}
Xiaobo Hu, Yang Liu, Yixin Ren, Yuan Gong

Contributors are listed alphabetically by their first name.

%% file: sections/appendix.tex
\appendix
\newpage

\section{Prompts}
\subsection{Synthetic Data Generation}
\label{app:synthetic_data_generation}
\begin{table*}[htbp]
\centering
\begin{minipage}{0.99\linewidth}\vspace{0mm}    
\centering
\begin{tcolorbox}[colframe=black!75!white, colback=white, coltitle=white, title=Prompt for Workflow Extraction (Step 1), fonttitle=\bfseries]
\begin{lstlisting}[basicstyle=\small\ttfamily]
You are an expert at extracting structured information from unstructured text. Given the following human-made question, extract the workflow steps and represent them in a structured JSON format.

Human Question: {task}

Analyze this question and break it down into logical workflow steps that would be needed to answer it completely.

Output format: return as **JSON**:
{
  "task": "{task}",
  "task_type": "{task_type}",
  "workflow": [
    {
      "step_id": 1,
      "description": "Description of step 1",
      "input": "Input details for step 1",
      "output": "Expected output for step 1",
      "dependencies": []
    },
    {
      "step_id": 2,
      "description": "Description of step 2",
      "input": "Input details for step 2",
      "output": "Expected output for step 2",
      "dependencies": [1]
    }
  ]
}

Instructions:
- Break down the task into logical, sequential steps
- Each step should have clear input/output requirements
- Include dependencies between steps where applicable
- Focus on actionable, concrete steps
- Ensure the workflow completely addresses the original question
\end{lstlisting}

\end{tcolorbox}
\caption{Prompt for Workflow Extraction.}

\end{minipage}
\end{table*}

\begin{table*}[htbp]
\centering
\begin{minipage}{0.99\linewidth}\vspace{0mm}    
\centering
\begin{tcolorbox}[colframe=black!75!white, colback=white, coltitle=white, title=Prompt for Attachment Searching (Step 2), fonttitle=\bfseries]
\begin{lstlisting}[basicstyle=\small\ttfamily]
You are an expert at generating Google Image search queries to find images that can be used as attachments for creating similar workflow-based questions across different scenarios.

Workflow Steps:
{workflow}

Your goal is to find images that:
1. Contain enough visual information to create questions that follow a similar workflow pattern
2. Are from realistic, practical scenarios (business documents, real interfaces, actual products, etc.)
3. Can be applied to different contexts while maintaining the same workflow structure
4. Have clear, readable details that an AI agent would need to analyze

Generate {num_queries} diverse and specific Google Image search queries that would find:
- Real-world examples with rich details (text, numbers, structures)
- Documents, interfaces, or visuals from different industries/domains
- Images that require similar analysis steps but in varied contexts

Guidelines for queries:
- Be specific about the type of document/interface/visual needed
- Include keywords that suggest detail-rich content ("detailed", "with text", "example", "template", "filled")
- Target professional or educational materials that contain substantive information
- Avoid generic or abstract queries
- Focus on finding images that contain actionable information

Examples of good queries:
- "detailed invoice template with line items filled out"
- "restaurant menu with prices and descriptions"
- "project gantt chart with tasks and deadlines"
- "ecommerce product page with specifications screenshot"
- "financial report dashboard with charts and metrics"
- "medical prescription form filled out example"
- "architectural floor plan with measurements and labels"

Return exactly {num_queries} queries as a JSON list.
Ensure queries are diverse across different domains but maintain similar complexity levels.

Queries:
\end{lstlisting}

\end{tcolorbox}
\caption{Prompt for Attachment Searching.}

\end{minipage}
\end{table*}

\begin{table*}[htbp]
\centering
\begin{minipage}{0.99\linewidth}\vspace{0mm}    \centering

\begin{tcolorbox}[colframe=black!75!white, colback=white, coltitle=white, title=Prompt for Query Generation (Step 3), fonttitle=\bfseries]
\begin{lstlisting}[basicstyle=\small\ttfamily]
You are an expert at generating diverse, high-quality questions based on workflow templates and question attachments. Given a workflow structure, generate new questions that follow the same logical pattern but with different content/context using attachments.

Original Workflow: {workflow}
Available Attachments: {available_attachments}
Attachments analysis: {attachment_analysis}

{task_type_description}

Generate {num_questions} new questions that follow this same workflow pattern. Each question should:
1. Follow the same logical workflow pattern
2. Have different content/domain/context than the original
3. Utilize attachments appropriately when provided
4. Have verifiable answers
5. Be realistic and meaningful
6. Matches the cognitive requirements of {task_type}
7. Be diverse in domain (tech, business, science, daily life, etc.)

Return as JSON:
{{
  "original_task": "{original_task}",
  "workflow_pattern": {workflow},
  "generated_questions": [
    {{
      "question_id": 1,
      "task_type": "{task_type}",
      "question": "New question following the workflow pattern",
      "domain": "Domain/context of this question",
      "attachments_used": ["List of attachment used"]
    }}
  ]
}}
\end{lstlisting}

\end{tcolorbox}
\caption{Prompt for Query Generation.}

\end{minipage}
\end{table*}

\begin{table*}[htbp]
\centering
\begin{minipage}{0.99\linewidth}\vspace{0mm}    
\centering
\begin{tcolorbox}[colframe=black!75!white, colback=white, coltitle=white, title=Prompt for Rubrics Generation (Step 4), fonttitle=\bfseries]
\begin{lstlisting}[basicstyle=\small\ttfamily]
Generate comprehensive evaluation criteria for the given task and workflow.
Task: {task}
Workflow: {workflow}

Create detailed evaluation rubric with bonus and penalty criteria following these principles:

**Scoring Guidelines:**
* **Bonus Criteria (+points)**: Represent completion of core assessment points of the task, such as:
  - Each node in the logical chain is a bonus item
  - Hidden conditions in "In context learning" are bonus items
  - Successfully achieving key task objectives

* **Penalty Criteria (-points)**: Include two categories:
  - Failure to meet basic requirements (e.g., incorrect output filename, wrong file type, etc.)
  - Making unnecessary and harmful modifications (e.g., changing content that should not be modified)

Return as JSON:
{{
  "evaluation_rubric": {{
    "task_id": "{task_id}",
    "bonus_criteria": [
      {{
        "description": "Specific core achievement to reward (logical chain node or key objective)",
        "points": 1,
        "target_output": "expected_output.txt or null",
        "step_id": 1,
        "verification": "How to verify this criterion",
        "category": "core_objective" // or "logical_chain_node" or "context_learning_condition"
      }}
    ],
    "penalty_criteria": [
      {{
        "description": "Basic requirement failure or harmful modification",
        "points": -1,
        "target_output": "incorrect_output.txt or null",
        "step_id": 2,
        "verification": "How to detect this error",
        "category": "basic_requirement_failure" // or "harmful_modification"
      }}
    ],
    "max_possible_score": 10,
    "min_possible_score": -5
  }}
}}

Requirements:
1. Each workflow step should have at least one evaluation criterion
2. Bonus criteria should focus on core task objectives and logical progression
3. Penalty criteria should clearly distinguish between basic requirement failures and harmful modifications
4. Points should be 1 or -1 primarily, occasionally 2 for critical core objectives
5. Criteria must be objectively verifiable
6. Include file format and naming convention checks in penalty criteria where applicable
7. Ensure penalty criteria specifically target unnecessary changes to protected content
\end{lstlisting}

\end{tcolorbox}
\caption{Prompt for Rubrics Generation.}

\end{minipage}
\end{table*}

\subsection{LLM as Judge}
\label{app:llm_as_judge}